\documentclass{article}

\usepackage[preprint,nonatbib]{neurips_2025}

\usepackage[utf8]{inputenc} 
\usepackage[T1]{fontenc}    

\usepackage[dvipsnames]{xcolor}
\definecolor{linkColor}{RGB}{237,4,140}
\definecolor{citecolor}{RGB}{0, 113, 188}
\usepackage[colorlinks=true,linkcolor=linkColor,citecolor=citecolor,filecolor=linkColor,urlcolor=linkColor]{hyperref}

\usepackage{url}            
\usepackage{booktabs}       
\usepackage{amsfonts}       
\usepackage{nicefrac}       
\usepackage{microtype}      

\usepackage{graphicx}
\usepackage{tabularx}
\usepackage{multirow}
\usepackage{booktabs}
\usepackage{bbding}
\usepackage{caption}
\usepackage{pifont}
\usepackage{stfloats}
\usepackage{bm}
\usepackage{wrapfig}
\usepackage{colortbl}
\usepackage{enumitem}
\usepackage{indentfirst}
\usepackage{algorithm}
\usepackage{algpseudocode}
\usepackage{amsmath}

\makeatletter
\newcommand\figcaption{\def\@captype{figure}\caption}

\title{DynamiCtrl: Rethinking the Basic Structure and the Role of Text for High-quality Human Image Animation}

\author{Haoyu Zhao$^{1}$,~Zhongang Qi$^{2}$,~Cong Wang$^{3}$,~Qingping Zheng$^{4}$,~Guansong Lu$^{2}$,~Fei Chen$^{2}$,\\~\textbf{Hang Xu}$^{2}$,~\textbf{Zuxuan Wu}$^{1}$ \\
\\
$^{1}$Fudan University, $^{2}$Huawei Noah’s Ark Lab, $^{3}$Sun Yat-sen University, $^{4}$Zhejiang University \\
\\
Code is available at \href{https://github.com/gulucaptain/DynamiCtrl}{https://github.com/gulucaptain/DynamiCtrl}.
\vspace{-0.3in}
}

\begin{document}

\maketitle

\begin{figure}[ht]
\begin{center}
	\includegraphics[width=0.90\linewidth]{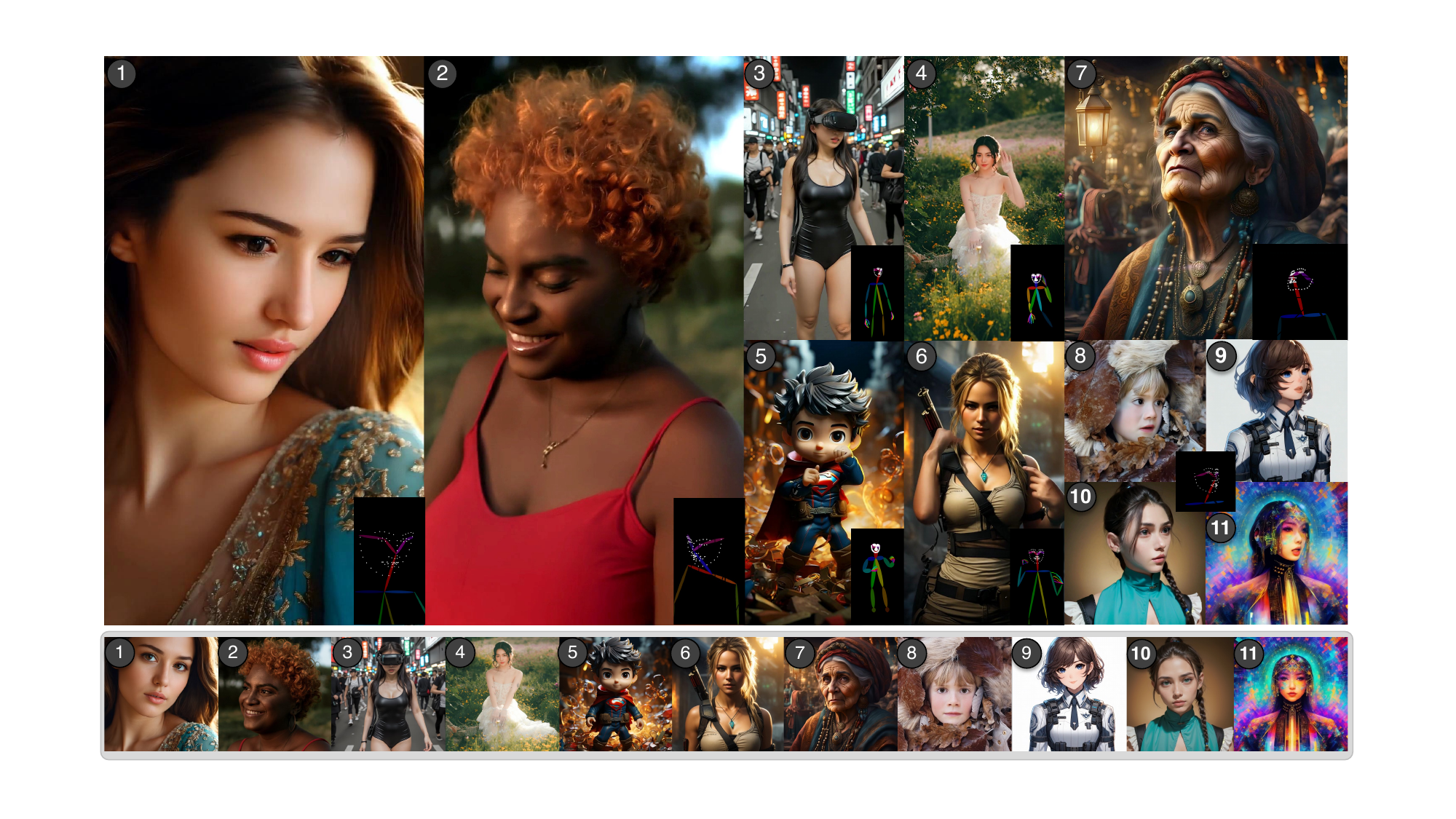}
\end{center}
\caption{\small
Pose-guided human image animation results by DynamiCtrl in 1360$\times$768 and 1360$\times$1360 resolutions. We show generated frames (top) with different persons (bottom), driven by precise pose sequence.
}
\vspace{-2pt}
\label{fig:teaser}
\end{figure}

\begin{abstract}
With diffusion transformer (DiT) excelling in video generation, its use in specific tasks has drawn increasing attention. 
However, adapting DiT for pose-guided human image animation faces two core challenges: (a) existing U-Net-based pose control methods may be suboptimal for the DiT backbone; and (b) removing text guidance, as in previous approaches, often leads to semantic loss and model degradation. To address these issues, we propose DynamiCtrl, a novel framework for human animation in video DiT architecture. Specifically, we use a shared VAE encoder for human images and driving poses, unifying them into a common latent space, maintaining pose fidelity, and eliminating the need for an expert pose encoder during video denoising. To integrate pose control into the DiT backbone effectively, we propose a novel Pose-adaptive Layer Norm model. It injects normalized pose features into the denoising process via conditioning on visual tokens, enabling seamless and scalable pose control across DiT blocks. Furthermore, to overcome the shortcomings of text removal, we introduce the ``Joint-text'' paradigm, which preserves the role of text embeddings to provide global semantic context. Through full-attention blocks, image and pose features are aligned with text features, enhancing semantic consistency, leveraging pretrained knowledge, and enabling multi-level control.
Experiments verify the superiority of DynamiCtrl on benchmark and self-collected data (\textit{e.g.,} achieving the best LPIPS of 0.166), demonstrating strong character control and high-quality synthesis.
\end{abstract}

\vspace{-2pt}
\section{Introduction} \label{sec:intro}
\vspace{-4pt}
With the rapid and magical growth the AIGC field is experiencing, pose-guided human image animation, as a task specifically for human motion video generation, has also gained increasing attention~\cite{yang2018pose,siarohin2021motion,chang2023magicpose,qin2023dancing,karras2023dreampose,ma2024follow}. The task aims to generate a video sequence in which a static human image is animated to follow a given pose sequence, with applications in social media, animation production, and digital human domains. Previous works~\cite{hu2024animate,zhang2024mimicmotion,zhu2024champ,wang2024loopanimate,xia2024musev,wang2024disco} have achieved promising results in this task by utilizing U-Net-based generative models~\cite{blattmann2023stable,chen2024videocrafter2,gu2023reuse,zhao2025magdiff} as the backbone.

Recently, represented by SD3~\cite{esser2024scaling} and Sora, the Diffusion Transformer (DiT) architecture has been widely used~\cite{peebles2023scalable, esser2024scaling, yang2024cogvideox, chen2024gentron, li2024hunyuan} and demonstrates superior quality in generating image and video contents. This also sparks exploration into how the driving pose can be implemented to achieve motion control within this new structure. In previous human animation methods~\cite{zhang2024mimicmotion,tu2024stableanimator,wang2024unianimate,tan2024animate}, the pose signal is typically combined with the noise in U-Net-based backbones~\cite{blattmann2023stable} after being processed by an expert pose encoder. However, due to the fundamental differences between U-Net and DiT, \textit{few works have thoroughly explored how to effectively integrate pose signals into the video DiT.}

In this work, we propose a novel framework for pose-guided human animation, named DynamiCtrl, which can generate animated videos with high quality, identity fidelity, and controllability (shown in Fig.~\ref{fig:teaser}). In our model, we first explore how to effectively control the pose signal in the video DiT architecture. Specifically, we used a \textbf{shared VAE encoder} to encode the training video, reference image, and raw pose sequence, instead of training an expert pose encoder as in previous works~\cite{wang2024disco,hu2024animate,zhu2024champ,zhang2024mimicmotion,tu2024stableanimator}, thereby ensuring the simplicity of our framework. Then, we propose the \textbf{Pose-adaptive Layer Norm (PadaLN)} to introduce the pose feature into the video DiT. In each DiT block, PadaLN employs learnable multilayer perceptrons to align the pose features with the statistical distribution of the noise. The patchified and normalized pose features are then added to the visual tokens, effectively coupling pose control with the denoising process and enabling precise pose-conditioned generation. In addition, we conduct an investigation of various pose control structures and highlight that directly integrating pose sequences into the DiT architecture, as adopted by previous methods~\cite{hu2024animate,zhang2024mimicmotion,tan2024animate} based on U-Net structures, is suboptimal for diffusion transformer frameworks.

Moreover, most previous methods~\cite{hu2024animate, moore, chang2023magicpose, zhu2024champ, wang2024unianimate, li2024dispose, zhang2024mimicmotion, tu2024stableanimator, peng2024controlnext} adopt a ``Non-text'' paradigm, removing the role of text from pre-trained text to image or video models~\cite{rombach2022high, blattmann2023stable}. A representative example is Animate Anyone~\cite{hu2024animate}, which replaces CLIP text features with corresponding image features, leveraging their shared embedding space to preserve semantic information. Later, subsequent approaches~\cite{zhang2024mimicmotion, tu2024stableanimator, peng2024controlnext} commonly use an image-based SVD model~\cite{blattmann2023stable} as the backbone, which is also pre-trained using the same CLIP image encoder. However, applying the ``Non-text'' paradigm to advanced video DiT models~\cite{yang2024cogvideox,li2024hunyuan,xu2024easyanimate} brings three key challenges: a) Increased training cost: Removing text from foundation models often requires additional training stages, making the process more resource-intensive; b) Loss of high-level semantics: existing video DiT models typically employ T5~\cite{raffel2020exploring} or MLLMs~\cite{sun2024hunyuan, glm2024chatglm} as text encoders, yet these models lack a directly aligned image space like CLIP~\cite{radford2021learning}, removing text on these backbones will lose high-level semantic information; c) Reduced controllability: losing the inherent flexibility offered by text, whereas combining textual prompts enables more versatile generation. So, we raise the question: \textit{is it possible to preserve the guidance of text within the DiT architecture, and what new horizons might this unlock for this specific task?}

With the question in mind, we refocus the role of text and shift from a ``Non-text'' to a \textbf{``Joint-text''} paradigm for human image animation. Specifically, during both training and inference, we minimize the impact on the backbone by preserving textual control while enabling pose control. After obtaining the unified visual tokens through PadaLN, we fuse them with textual tokens using a full-attention mechanism within each DiT block, facilitating effective multi-modal integration. Furthermore, by introducing masks into the human image, the ``Joint-text'' paradigm enables masked training without modifying the model architecture or introducing additional trainable parameters, thereby allowing fine-grained control over local details in the generated video. We summarize the \textbf{advantages} of the ``Joint-text'' paradigm as follows: a) One-time training: avoids additional training costs to remove the text signal; b) Preservation of textual semantic representations: helps to maintain pre-trained knowledge; c) Multi-level controllability: enables control not only over precise human motion, but also over customized video content.
In Table~\ref{tab:system_level_comparison}, we conduct a system-level comparison between our model and existing methods.
Based on the proposed ``Joint-text'' paradigm, we develop DynamiCtrl on two pre-trained models with 2B and 5B parameters. Our model is capable of generating $1360 \times 1360$ resolution videos with superior visual quality.
In short, our work makes the following contributions:

\begin{table*}[t!]
    \centering
    \small
    \caption{A system-level comparison between the proposed DynamiCtrl framework and existing State-of-the-Art human image animation frameworks.}
    \resizebox{\linewidth}{!}{
    \begin{tabular}{l|c|cc|ccc|c}
       \toprule
       \multirow{2}{*}{\textbf{Methods}} & \textbf{Generation} & \multicolumn{2}{c|}{\textbf{Performances}} & \multicolumn{3}{c|}{\textbf{Control Conditions}}  & \textbf{W/o Expert} \\
       \cmidrule{3-4} \cmidrule{5-7}
       & \textbf{paradigm} & \textbf{Resolution} & \textbf{LPIPS $\downarrow$} & \textbf{Human image} & \textbf{Pose} & \textbf{Text}  & \textbf{Pose-encoder} \\
       \cmidrule{1-1} \cmidrule{2-8}
       AnimateAnyone~\cite{hu2024animate} & Non-text & $784 \times 512$ & 0.285 & \Checkmark & \Checkmark & \XSolidBrush  & \XSolidBrush  \\
       Champ~\cite{zhu2024champ} & Non-text & $512 \times 512$ & 0.231 & \Checkmark & \Checkmark & \XSolidBrush  & \XSolidBrush  \\
       MimicMotion~\cite{zhang2024mimicmotion} & Non-text & $1024 \times 576$ & 0.414 & \Checkmark & \Checkmark & \XSolidBrush  & \XSolidBrush  \\
       StableAnimator~\cite{tu2024stableanimator} & Non-text & $1024 \times 576$ & 0.232 & \Checkmark & \Checkmark & \XSolidBrush  & \XSolidBrush \\
       Animate-X~\cite{tan2024animate} & Non-text & $768 \times 512$ & 0.232 & \Checkmark & \Checkmark & \XSolidBrush  & \XSolidBrush \\
       \midrule
       \textbf{DynamiCtrl} & \textbf{Joint-text} & \bm{$1360 \times 1360$} & \textbf{0.166} & \Checkmark & \Checkmark & \Checkmark  & \Checkmark \\
       \bottomrule
    \end{tabular}}
    \vspace{-10pt}
    \label{tab:system_level_comparison}
\end{table*}

\begin{enumerate}
[topsep=3.5pt,itemsep=3pt,leftmargin=20pt]
\item We propose a novel DynamiCtrl framework for pose-guided human animation, which is based on the video DiT architecture. We employ a shared VAE encoder for both human images and driving poses, and design a Pose-adaptive Layer Normalization (PadaLN) model to effectively enable precise human motion control.
\item We propose a ``Joint-text'' paradigm for human image animation that preserves textual control alongside pose guidance, enabling fine-tuning with high-level semantic features, retaining strong pre-trained priors, and achieving fine-grained controllability with one-time training.
\item Experimental results on benchmark and self-collected datasets demonstrate that DynamiCtrl achieves accurate motion control and improved quantitative performances, \textit{e.g.,} best LPIPS result of 0.166, along with high-quality in both fine-grained visual details and identity preservation.
\end{enumerate}

\vspace{-2pt}
\section{Related Work} \label{sec:related}
\vspace{-2pt}
\subsection{Image and Video Generation}
Diffusion Models~\cite{2020DMs,blattmann2023stable,peebles2023scalable,gu2023reuse,yang2024cogvideox,zhao2025magdiff,zhang2025eden} have recently demonstrated impressive results in image and video synthesis.
Earlier, U-Net-based models demonstrated satisfactory performance in text control and visual expression. 
Image diffusion models~\cite{rombach2022high} aim to learn to generate high-quality images through a process of progressive denoising, while video diffusion models~\cite{blattmann2023stable} focus on achieving good temporal consistency.
Methods~\cite{chen2024videocrafter2, wang2023videocomposer, zhao2025magdiff, TI2V-Zero} rely on temporal layers or 3D structure to keep temporal consistency.
Later, with the rise of the Sora, models based on the video DiT architecture gradually became mainstream, showing stronger performance in terms of generation quality, including both image~\cite{peebles2023scalable,li2024hunyuan} and video~\cite{xu2024easyanimate,yang2024cogvideox} generation.

\subsection{Human Image Animation}
Human image animation refers to the process of creating a dynamic video that depicts human motion based on pose sequences, typically generated by diffusion models.
In the early stage, image-diffusion-based methods~\cite{siarohin2021motion, qin2023dancing, karras2023dreampose, ma2024follow, wang2024loopanimate, wang2024disco, xia2024musev, hu2024animate} typically rely on post-processing to generate human motion video, which often leads to loss of detail and heavy temporal inconsistency. 
Subsequently, video-diffusion-based methods~\cite{chang2023magicpose, xu2024magicanimate, zhang2024mimicmotion} alleviate the problem of temporal inconsistency, relying on the capabilities of the backbone model. 
However, since U-Net lags behind the DiT architecture~\cite{peebles2023scalable,chen2024gentron}, these models still face significant challenges in maintaining character identity.
Although Methods~\cite{luo2025dreamactor, lin2025omnihuman} attempt pose-guided video control on the DiT, they still rely heavily on direct adaptation of U-Net-based structures without fully exploring pose control designs.
Additionally, the lack of text input in existing methods prevents us from further controlling the generated content.

\vspace{-2pt}
\section{Method} \label{sec:method}
\vspace{-2pt}
\begin{figure*}[ht]
    \centering
    \captionsetup{type=figure}
    \includegraphics[width=1.0\linewidth]{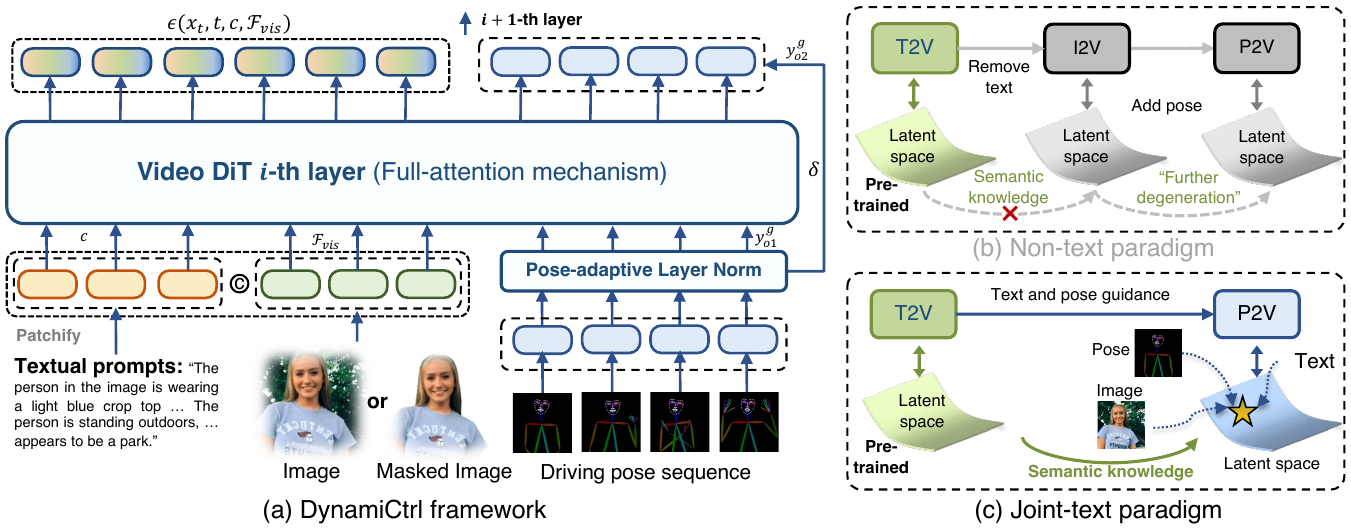}
    \caption{Overview of the proposed DynamiCtrl framework for human image animation. In (b) and (c), the T2V, I2V, and P2V denote the text-guided, image-guided, and pose-guided video generation.}
    \vspace{-10pt}
    \label{fig:framework}
\end{figure*}

Our goal is to advance human image animation by incorporating effective pose-guided motion control, while preserving the powerful generative capabilities of diffusion transformer-based pre-trained models.
To this end, we present the DynamiCtrl framework (see Fig.~\ref{fig:framework}), which consists of a shared VAE encoder, a Pose-adaptive Layer Norm that injects pose features into the video DiT, several full-attention blocks for denoising, and a novel ``Joint-text'' paradigm for training and inference.

\subsection{DynamiCtrl Framework}
\label{sec:_dynamictrl_framework}
Given a human reference image $I$ and a driving pose video $I_{1:F}$, before entering the video DiT blocks, we first consider how to extract features from the image and the pose video, as well as how to organize them together within the DiT architecture.

\paragraph{Shared VAE Encoder.}
Following a typical image preserving structure~\cite{zhao2025magdiff,yang2024cogvideox}, we use a VAE encoder $\mathcal{E}$ to extract features $\mathcal{F}_i$ from the human image, which is also responsible for mapping the training video from pixel space to latent space.
We concatenate the $\mathcal{F}_i$ with the noise latent $\mathcal{N}$ as the vision latent.
For pose driving video $I_{1:F}=\{I_1,I_2,...,I_F\}$, almost all the previous methods~\cite{hu2024animate,wang2024disco,zhang2024mimicmotion,tan2024animate} need to train an expert pose encoder to encode the driving video.
For instance, methods~\cite{hu2024animate,zhang2024mimicmotion} use shallow convolution layers as the pose guider to encode the pose sequence, and they initialize the module with Gaussian weights. Disco~\cite{wang2024disco} and MagicAnimate~\cite{xu2024magicanimate} adopt structures like ControlNet~\cite{controlnet} to introduce the pose information into the diffusion process. 
Although the effectiveness of the expert pose encoder has been validated, the addition of a new encoder not only increases the complexity of the model architecture but also introduces additional training burdens, especially in designs like ControlNet, where the parameter count can rise from 300M to over 1B.

In our design, considering that the video DiT block will directly process the visual features after unified encoding, we attempt to reuse the VAE encoder $\mathcal{E}$ to encode the driving pose video into the latent space.
The raw driving video (for $f \times 3 \times h \times w$) is permuted to $3 \times f \times h \times w$ and handled by VAE encoder $\mathcal{E}(I_{1:F})$ to $\mathcal{F}_v \in \mathbb{R}^{F \times C \times H \times W}$. The $\mathcal{F}_v$ is also multiplied by the scaling factor $\kappa$ of the VAE.
This design eliminates the need for expert pose encoders, thereby simplifying the architecture, and we don't retrain the VAE encoder. Furthermore, it ensures that the features of these different inputs are aligned in the same latent embedding space, making it more suitable for the video DiT model that requires a unified input of visual tokens.

\paragraph{Bringing Pose Control into Video DiT.}
After the shared VAE encoder, we get the features $\mathcal{F}_v$ of driving video. To inject the feature into video DiT blocks with reference image features $\mathcal{F}_i$ and noise latent $\mathcal{N}$, we explore three variants of model structures that process pose features differently.

\begin{itemize}
[topsep=3.5pt,itemsep=3pt,leftmargin=10pt]
\item[-] \textbf{Unify Vision Token (UVT)}. The image features $\mathcal{F}_i$ are concatenated with latent $\mathcal{N}$ with zero padding. The output channel after the VAE encoder is $C$, so the channel number becomes $2 \times C$. The most straightforward way is to continue concatenating the pose features with these vision tokens, treating $\mathcal{F}_v$ no differently than the $\mathcal{F}_i$ and the $\mathcal{N}$. In this way, the channel numbers of latent changes to $2 \times C + C$. In video DiT blocks, the unified vision tokens are all used as the query ($Q$), key ($K$), and value ($V$) in full attention layers.

\item[-] \textbf{Cross Attention}.
In previous video generation works~\cite{chen2024gentron, li2024hunyuan, zhao2025magdiff, gu2023reuse}, the Cross-Attention (CA) is a widely used structure to fuse the text or image signals with vision features.
Following the same design, we try to inject $\mathcal{F}_v$ using cross-attention. Using feature patchy, we translate feature $\mathcal{F}_v$ to hidden representation $y^g \in \mathbb{R}^{B \times T \times D} $. After full attention layer, we get the output $\hat{y}^g \in \mathbb{R}^{B \times T \times D} $ of $\mathcal{F}_i$ and $\mathcal{N}$.
In cross-attention, we represents the query ($Q$), key ($K$), and value ($V$) as follows:

\begin{equation}
\label{eq: eq1}
[Q, K, V] = [W^{Q}\hat{y}^{g},W^{K}y^{g},W^{V}y^{g}]
\end{equation}

After the cross-attention layer, we add the $\hat{y}^g$ and the output together. Under this design, we observe that the effectiveness of pose control is not satisfactory. It fails to effectively capture the nonlinear relationship between pose and generated videos, leading to a lack of clear control. 
Besides, unfortunately, it also brings a significant increase in parameters, adding up to 40\% more.

\item[-] \textbf{Pose-adaptive Layer Norm (PadaLN)}.
Considering the difference between the U-Net and DiT structures, we further propose the Pose-adaptive Layer Norm (PadaLN), which utilizes normalization to integrate pose features $\mathcal{F}_v$ into DiT while preserving the model's spatiotemporal relationship modeling capacity. After getting the hidden representation $y^g$ of $\mathcal{F}_v$, leveraging the MLP, we learn the scale $\beta$, shift $\gamma$, and gate $\delta$ parameters from the embedding vectors of time $t$:

\begin{equation}
\label{eq: eq2}
\gamma,\beta,\delta=MLP(SiLU(t))
\end{equation}

Before the full attention, we use $\beta$ and $\gamma$ to rescale the representation $y^g$ as follows:

\begin{equation}
\label{eq: eq3}
y^{g}_{o1}=LN(y^{g})*(1 + \beta) + \gamma,
\end{equation}

where $LN$ denotes the normalization layer. We add the $y^{g}_{o1}$ to the hidden representation of the noise latent. We do not adopt the concatenation method because the element-wise summation operation preserves channel dimensions, thus keeping the model structure unchanged. Besides, before the PadaLN, we repeat the pose features along the channel dimension to keep the dimensions consistent. The scaled feature $y^{g}_{o1}$ is calculated by $\delta$ to get the input $y^{g}_{o2}$ for next video DiT block:

\begin{equation}
\label{eq: eq4}
y^{g}_{o2} = y^{g}_{o1} + \delta \times y^{g}_{o1}
\end{equation}
\end{itemize}

\begin{wrapfigure}[18]{r}{0.49\textwidth}
\centering
\vspace{-12pt}
\includegraphics[width=0.45\textwidth]{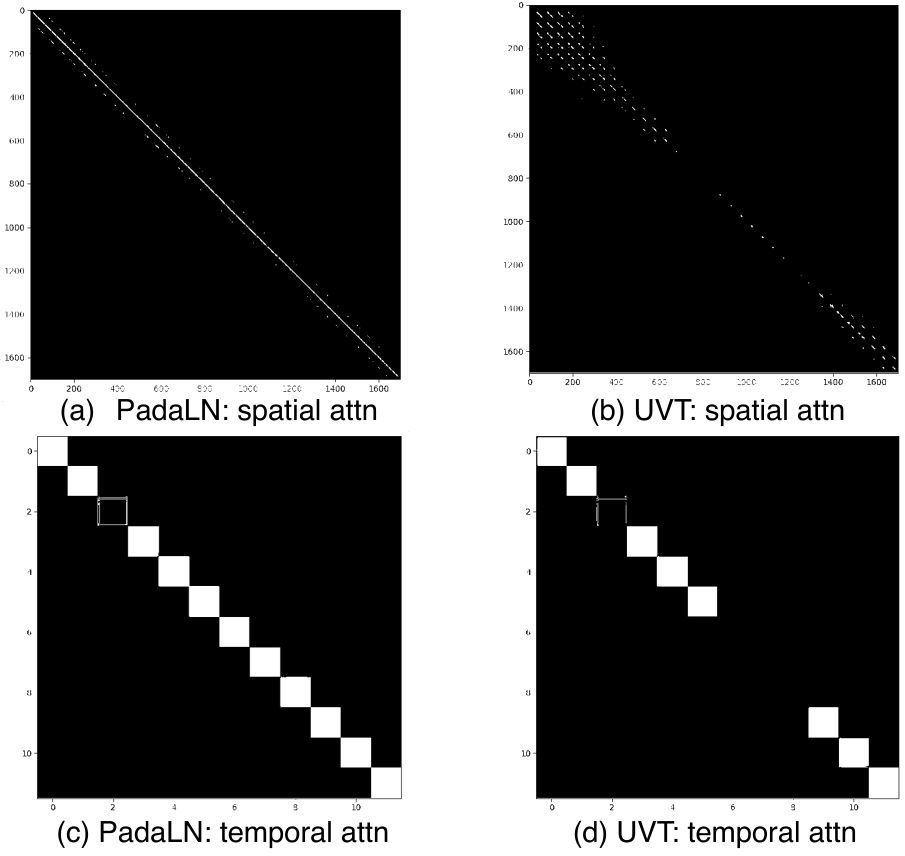}
\vspace{-5pt}
\caption{\small
Vision-to-vision spatial and temporal attention visualizations of different control methods.
}
\label{fig:attention_visualization}
\end{wrapfigure}
\vspace{-3pt}

Among these structures, our ablations in Table~\ref{tab: pose_control_comparison} show that the PadaLN in DynamiCtrl achieves the best performance.
\textit{Intuitively, PadaLN normalizes the $\mathcal{F}_v$ to match the statistical distribution of the noise $\mathcal{N}$, thereby facilitating more effective guidance of the generation trajectory within the DiT model.} It ensures that the pose signal can influence the generation process without disrupting the inherent noise distribution.
Furthermore, following the visualization method~\cite{cai2024ditctrl}, we illustrate the binarized attention maps of DiT with UVT and PadaLN in Fig.~\ref{fig:attention_visualization}, showing spatial attention at the single-frame level and temporal attention by tracking tokens at identical spatial positions across different frames.
Notably, in PadaLN, we observe \textbf{a distinctive diagonal pattern} in both spatial and temporal attention, indicating strong spatial-temporal correlations and close dependencies between adjacent video frames.
On the contrary, in UVT, the diagonal pattern is missing in some areas, indicating that the model struggles to capture frame relationships, especially in long-term sequences.

\subsection{``Joint-text'' Paradigm for Human Image Animation}
For the human image animation task, previous works~\cite{hu2024animate,xu2024magicanimate,zhang2024mimicmotion,tan2024animate} mainly follow a ``Non-text'' paradigm, in which textual input is typically omitted.
Although the ``Non-text'' paradigm is effective in practice, it presents several critical issues:
a) \textit{Increased training cost:} since almost all pre-trained models~\cite{yang2024cogvideox,sun2024hunyuan} are conditioned on text, additional multi-stage training is required to remove the text input before incorporating motion control signals;
b) \textit{Loss of high-level semantics:} video DiT models typically employ T5~\cite{raffel2020exploring} or MLLMs~\cite{sun2024hunyuan, glm2024chatglm} as text encoders instead of the CLIP model~\cite{radford2021learning}, which leads to the absence of an aligned image encoder. Following the ``Non-text'' paradigm, we have to remove the original text inputs entirely. However, due to the lack of high-level semantic guidance during training, the generation ability and the pretrained knowledge of the fine-tuned animation model are degraded;
c) \textit{Reduced controllability:} The resulting animation model completely loses the ability to flexibly control generation through textual prompts.

In this work, we propose a new \textbf{``Joint-Text'' paradigm} for human image animation, which explores how to achieve precise motion control within advanced video DiT backbones. We emphasize that, unlike methods~\cite{luo2025dreamactor,lin2025omnihuman,hu2024animate,zhang2024mimicmotion,tan2024animate} which remove text from the DiT or U-Net backbones, the proposed ``Joint-text'' paradigm preserves the text modality during training and inference. By leveraging the high-level semantic features provided by text, we enable better control over the generated results.

\textbf{Preservation of textual semantic representations.} Formally, we define the conditions to be a tuple $\mathcal{G}= \left \langle I,I_{1:F},\mathcal{C} \right \rangle $, where $\mathcal{C}$ denotes the space of textual descriptions. We obtain the text feature $c$ from text encoder $Emb(\mathcal{C})$.
After PadaLN, unified visual tokens $\mathcal{F}_{\mathrm{vis}}$ are added with the text tokens $c$, fed into the DiT.
Intuitively, the video generator $\rho$ synthesizes $K$-step video trajectories that illustrate possible diffusion processes for completing the human animation task with $\{c,\mathcal{F}_{\mathrm{vis}}\}$.
In the reverse diffusion, we aim to restore the video using a denoising network $\epsilon_\theta$. The prediction paradigm is:

\begin{equation}
\epsilon_{\theta}(x_{t},t,\rho)=\epsilon_{\theta}(x_{t},t,c,\mathcal{F}_{\mathrm{vis}})
\end{equation}

\noindent The whole reverse diffusion process to generate videos with $\{c,\mathcal{F}_{\mathrm{vis}}\}$ can be represented as Eq.~\ref{eq:six}, where $z\sim\mathcal{N}(0,\mathbf{I})$ is the standard Gaussian noise and $\sigma_{t}$ is the deviation related to timestep $t$. By encoding text into latent representations, we can apply diffusion to high-level semantic structures, instead of low-level pixel couplings between the reference image and given pose.

\begin{equation}
\label{eq:six}
x_{t-1}=\frac{1}{\sqrt{\alpha_t}}\left(x_t-\frac{1-\alpha_t}{\sqrt{1-\bar{\alpha}_t}}\epsilon_\theta(x_t,t,c,\mathcal{F}_{\mathrm{vis}})\right)+\sigma_tz
\end{equation}

\textbf{One-time training.} While retaining textual control, we further provide high-level semantic features during finetuning to prevent the pretrained model from degrading into low-level pixel space representations. The whole loss for training with text input of our model is designed as:

\begin{equation}
L=\mathbb{E}_{x_0,t,\epsilon}\left[\left\|\epsilon-\epsilon_\theta(x_t,t,c,\mathcal{F}_{\mathrm{vis}})\right\|^2\right],
\end{equation}

\noindent where the $\epsilon$ is the added noise. So, we establish the connection between visual and textual signals, enabling the model to be simultaneously guided by human images, driving videos, and prompts. As shown in Fig.~\ref{fig:framework} (b) and (c), our ``Joint-text'' paradigm eliminates the need for an intermediate training and enables the animation model to be trained in one-time training, effectively preserving the pretrained knowledge through full-attention alignment between visual and textual tokens.

\textbf{Multi-level controllability.} 
Moreover, thanks to the preservation of textual semantic features, our framework can be naturally extended to multi-level controllable generation. To fully leverage the guidance provided by text, we apply a mask to the image $I$ to obtain the masked visual tokens $\mathcal{F}_{\mathrm{vis}}^{\mathcal{M}}$, which are then aligned with the textual tokens within the full-attention modules. As shown in Fig.~\ref{fig:background_control}, this approach enables fine-grained control over local regions in the video without modifying the model architecture or introducing additional training parameters.

\vspace{-2pt}
\section{Experiments} \label{sec:experiments}
\vspace{-2pt}
\subsection{Implementation Details}

We build the DynamiCtrl model upon the pre-trained CogVideoX~\cite{yang2024cogvideox}, which comes in two variants with different parameter scales. Specifically, we refer to them as ``DynamiCtrl-2B'' and ``DynamiCtrl-5B'', containing 2 billion and 5 billion parameters, respectively. During training, we employ a supervised fine-tuning (SFT) strategy.

\textbf{Evaluations.} To evaluate our method, following previous works~\cite{wang2024disco,hu2024animate,zhang2024mimicmotion,tu2024stableanimator,zhu2024champ,chang2023magicpose}, we use the popular TikTok benchmark~\cite{jafarian2021learning}, which contains 350 videos. We train our method on the training set (1-334) and test on the test set (335-340). Additionally, since the training data of open-source models in previous methods are all private due to human privacy policies, we collect about 13k human motion videos from the internet for training and 100 unseen videos, referred to as Unseen100, as the test set to conduct zero-shot inference.

\paragraph{Baseline.}
Although existing methods~\cite{luo2025dreamactor, lin2025omnihuman} also perform human animation based on the DiT architecture, they are all closed-source. Therefore, we build our baseline on the same pre-trained CogVideoX model for fair comparison, named ``Baseine-2B / 5B''. We follow methods~\cite{hu2024animate,zhang2024mimicmotion} by removing the text input, employing an expert pose encoder for the driving video, and directly adding the pose features into the visual tokens.

\begin{table*}[t]
    \centering
    \caption{Quantitative comparisons with SOTAs on TikTok and Unseen100. ``PSNR*''~\cite{wang2024unianimate} denotes the modified metric to prevent numerical overflow. In the table, ``a / b'', a, and b denote results on TikTok and Unseen100, respectively. ``-'' means the missing values in their papers.}
    \vspace{-5pt}
    \resizebox{1.0\linewidth}{!}{
    \begin{tabular}{l|ccccccc|cc}
       \toprule
       \multirow{2}{*}{\textbf{Methods}} && \multicolumn{5}{c}{\textbf{Frame Quality}} && \multicolumn{2}{c}{\textbf{Video Quality}} \\
       \cmidrule{3-7} \cmidrule{9-10}
       && L1 (E-04) $\downarrow$ & SSIM $\uparrow$ & PSNR $\uparrow$ & PSNR* $\uparrow$ & LPIPS $\downarrow$ && FID-VID $\downarrow$ & FVD $\downarrow$ \\
       \midrule
       MagicAnimate~\cite{xu2024magicanimate} && 3.13 / 8.21 & 0.714 / 0.351 & 29.16 / 21.13 & - / 11.44 & 0.239/ 0.513 && 21.75 / 85.61 & 179.07 / 1058.37 \\
       Animate Anyone~\cite{hu2024animate} && - / 3.46 & 0.718 / 0.572 & 29.56 / 26.60 & - / 16.65 & 0.285 / 0.283 && - / 39.26 & 171.90 / 723.93 \\
       Champ~\cite{zhu2024champ} && 2.94 / 4.23 & 0.802 / 52.39 & 29.91 / 25.91 & - / 15.78 & 0.231 / 0.319 && 21.07 / 25.48 & 160.82 / 535.96 \\
       Unianimate~\cite{wang2024unianimate} && 2.66 / 4.46 & \textbf{0.811} / 0.503 & 30.77 / 24.34 & 20.58 / 14.61 & 0.231 / 0.344 && - / 33.39 & 148.06 / 837.93 \\
       MimicMotion~\cite{zhang2024mimicmotion} && 5.85 / 3.60 & 0.601 / 0.618 & - / 27.39 & 14.44 / 17.09 & 0.414 / 0.248 && - / 23.04 & 232.95 / 417.56 \\
       ControlNext~\cite{peng2024controlnext} && 6.20 / 3.53  & 0.615 / 0.577 & - / 27.89 & 13.83 / 17.20 & 0.416 / 0.263 && - / 21.16 & 326.57 / 414.22 \\
       StableAnimator~\cite{tu2024stableanimator} && 2.87 / 7.86 & 0.801 / 0.448 & \textbf{30.81} / 23.11 & 20.66 / 12.00 & 0.232 / 0.423 &&  - / 22.87 & 140.62 / 365.46 \\
       Animate-X~\cite{tan2024animate} && 2.70 / 3.38 & 0.806 / 0.619 & 30.78 / 27.42 & \textbf{20.77} / 17.80 & 0.232 / 0.259 &&  - / 26.95 & \textbf{139.01} / 570.98 \\
       \midrule
       Baseline-2B && 7.22 / 7.09 & 0.756 / 0.597 & 28.43 / 26.32 & 18.60 / 16.21 & 0.223 / 0.269 && 18.99 / 22.21 & 176.47 / 287.34 \\
       \rowcolor{lightgray!20}
       DynamiCtrl-2B && 5.57 / 5.72 & 0.773 / 0.616 & 29.91 / 27.90 & 19.30 / 17.78 & 0.201 / 0.249 && 17.90 / 20.57 & 172.31 / 276.09 \\
       Baseline-5B && 4.41 / 4.69 & 0.745 / 0.618 & 29.31 / 27.84 & 19.54 / 17.38 & 0.192 / 0.251 && 14.73 / 19.30 & 163.45 / 289.11 \\
       \rowcolor{lightgray!20}
       \textbf{DynamiCtrl-5B} && \textbf{2.34} / \textbf{3.09} & 0.766 / \textbf{0.623} & 30.22 / \textbf{29.43} & 20.41 / \textbf{18.37} & \textbf{0.166} / \textbf{0.235} && \textbf{13.77} / \textbf{17.15} & 152.31 / \textbf{249.80} \\
       \bottomrule
    \end{tabular}}
    \vspace{-10pt}
    \label{tab:evaluation_on_tiktok}
\end{table*}
\vspace{-5pt}

\subsection{Comparison with State-of-the-Art Methods}
\label{sec:comparison_with_methods}

\textbf{Quantitative results.}
In Table~\ref{tab:evaluation_on_tiktok}, we compare the quantitative results with state-of-the-art methods~\cite{zhu2024champ,wang2024unianimate,zhang2024mimicmotion,peng2024controlnext,tu2024stableanimator,tan2024animate} on the TikTok and Unseen100 datasets.
On TikTok, our method achieves the best performance on the L1, LPIPS, and FID-VID metrics, while also delivering comparable results on the PSNR and FVD metrics. Our LPIPS score gets 0.166, representing \textbf{a 28.1\% improvement} over the best-performing method~\cite{zhu2024champ}.
To obtain the results of the compared methods, we directly use their open-source models.
Since methods~\cite{lin2025omnihuman,luo2025dreamactor} based on DiT are closed-source and have not been evaluated on public datasets, we are unable to perform a direct comparison of results.

\textbf{Compared with Baseline model.}
Besides, in Table~\ref{tab:evaluation_on_tiktok}, we further compare the performance of DynamiCtrl with the baseline. The results show that, under the same size of parameters in the same pre-trained model, 2B or 5B model, we outperform the baseline or achieve competitive performances. 
This demonstrates the effectiveness of our motion control model and text-driven paradigm, with improvements not solely due to a stronger backbone.

\vspace{-5pt}
\begin{figure*}[ht]
    \centering
    \captionsetup{type=figure}
    \includegraphics[width=0.95\linewidth]{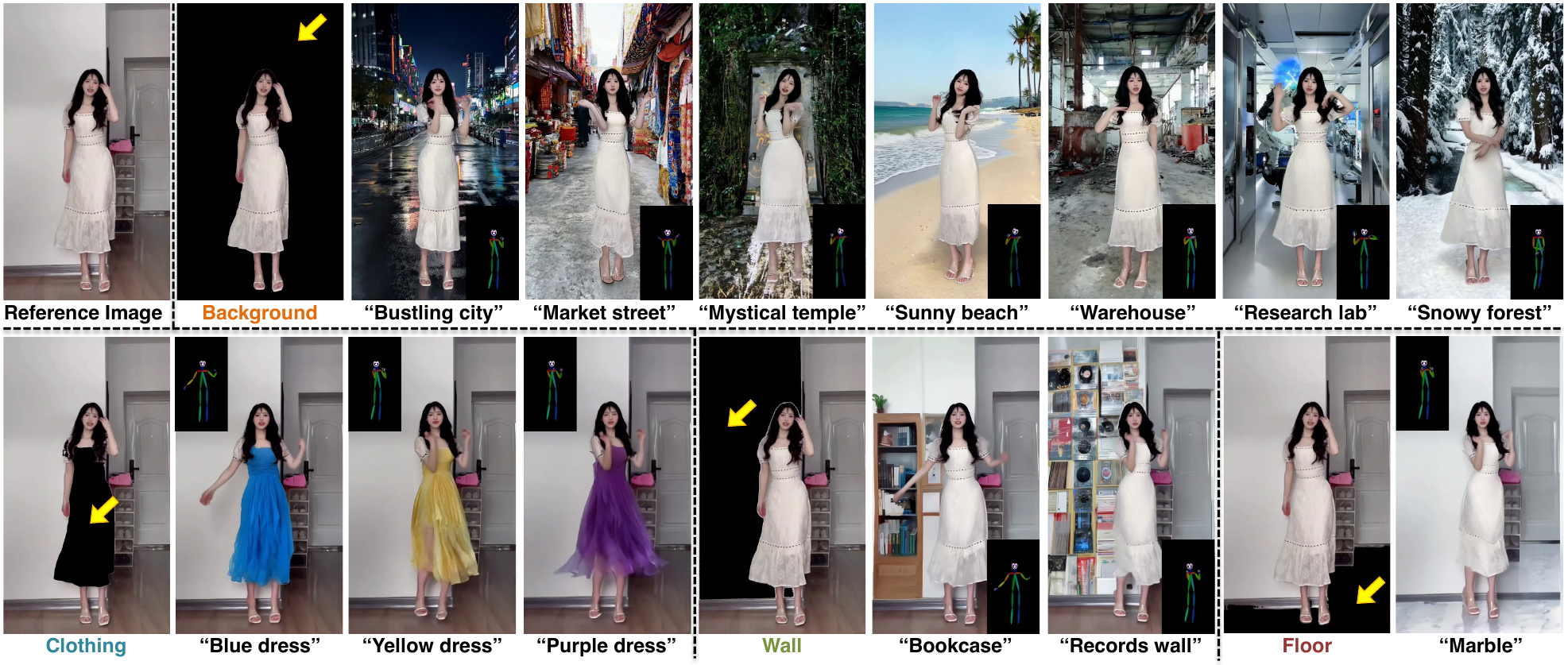}
    \caption{The ``Joint-text'' paradigm enables our model to achieve multi-level controllability, allowing not only precise human motion, but also fine-grained control over all elements in the image.}
    \label{fig:background_control}
\end{figure*}

\textbf{Qualitative results: multi-level controllability.}
The proposed ``Joint-text'' paradigm is specifically designed for the human image animation task, addressing several limitations encountered by previous ``Non-text'' paradigm. As illustrated in Fig.~\ref{fig:background_control}, incorporating masked regions of the image during training and inference allows our method to control both human motion and other image elements through text. As a result, the model trained for this specific task retains strong capabilities in text-driven control and open-ended generation, thereby expanding the range of practical applications.

\begin{figure*}[!t]
    \centering
    \captionsetup{type=figure}
    \includegraphics[width=0.95\linewidth]{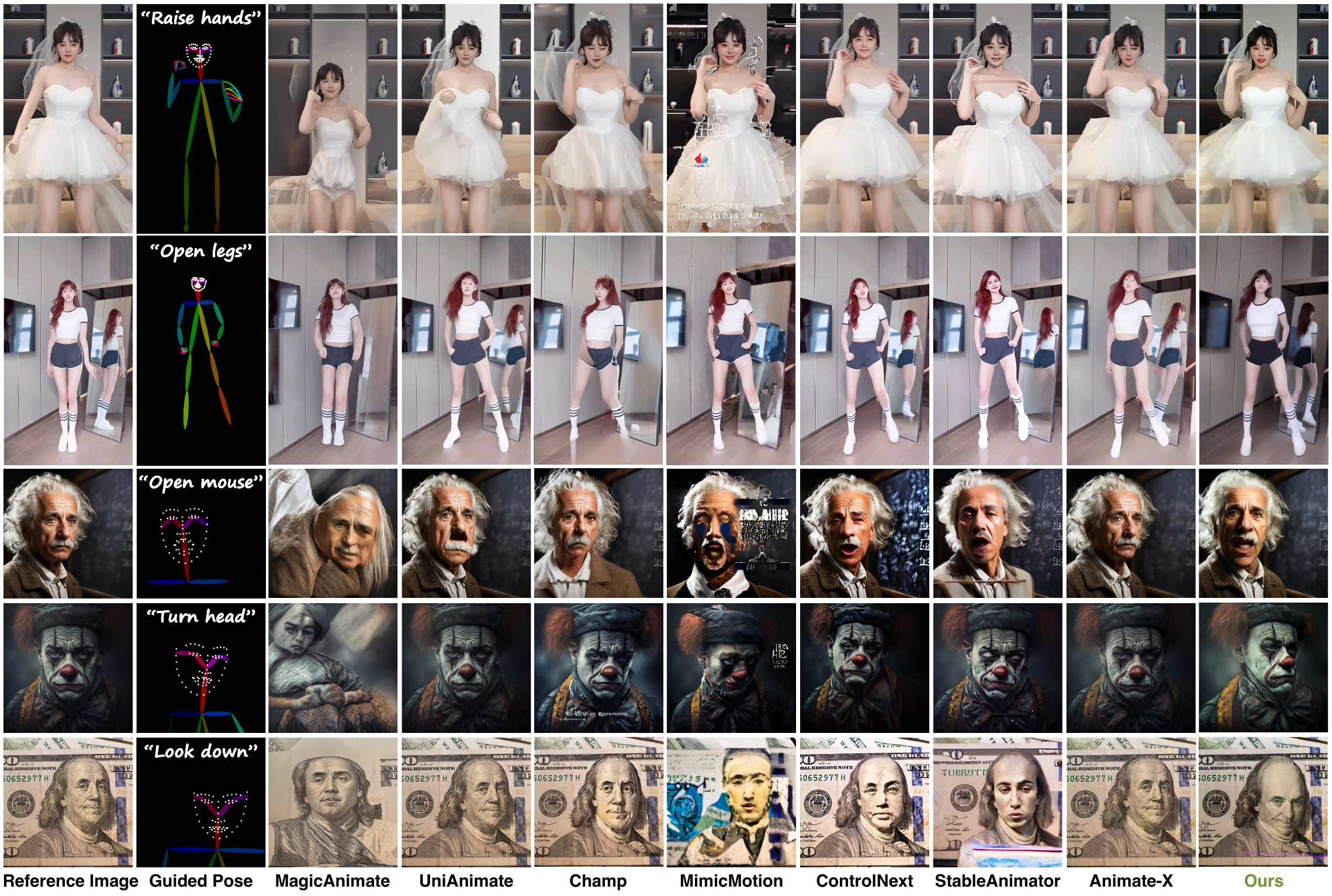}
    \caption{Qualitative comparisons with SOTAs on five challenging unseen examples. We use their released models for generation, and Animate-X~\cite{tan2024animate} is the latest open-source animation model.}
    \vspace{-10pt}
    \label{fig:visualization}
\end{figure*}
\vspace{-5pt}

\begin{figure*}[ht!]
    \centering
    \captionsetup{type=figure}
    \includegraphics[width=1.0\linewidth]{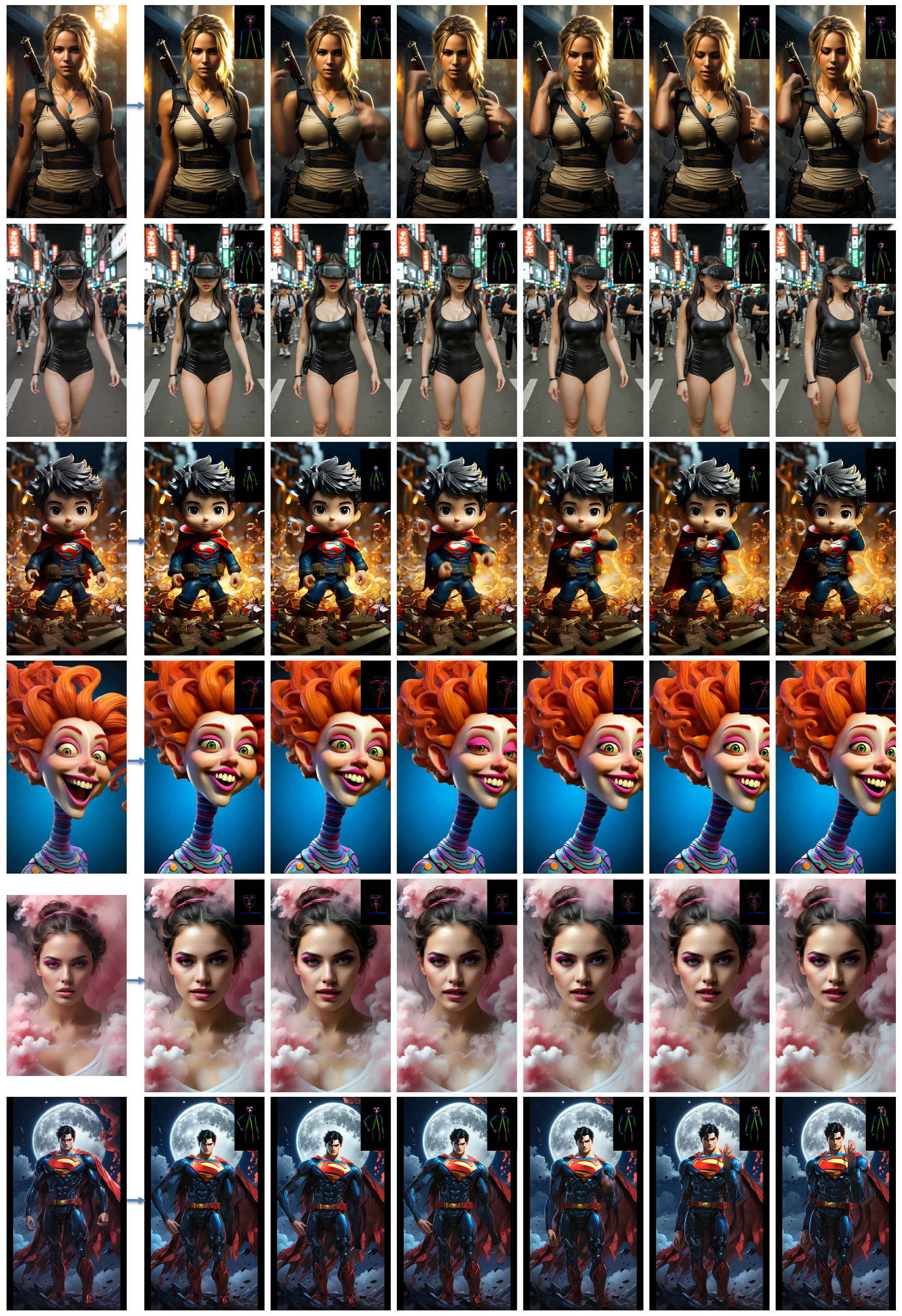}
    \caption{Human Image Animation Performance. We present examples with resolutions up to 1360 × 768. The reference images are shown on the leftmost side. Specifically, our model can generate motion for anime characters while preserving facial fidelity.}
    \label{fig:more_visualization}
\end{figure*}

\begin{figure*}[ht!]
    \centering
    \captionsetup{type=figure}
    \includegraphics[width=0.92\linewidth]{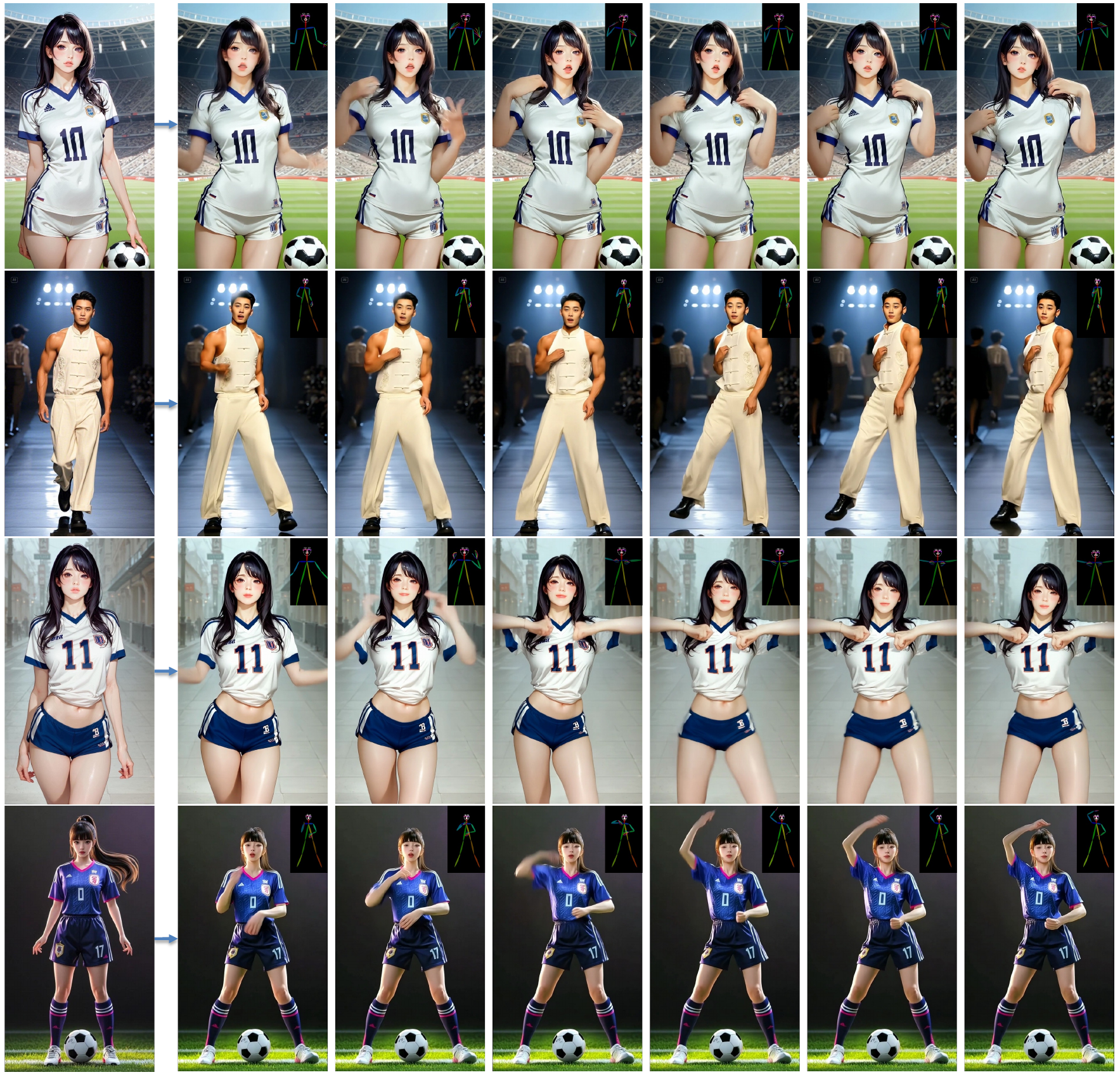}
    \caption{Performance in following large motions with jazz dance.}
    \label{fig:large_motion_cases}
\end{figure*}

\paragraph{Qualitative results: compared with other models.}
In Fig.~\ref{fig:visualization}, we provide qualitative comparisons with six state-of-the-art methods~\cite{xu2024magicanimate,wang2024unianimate,zhang2024mimicmotion,peng2024controlnext,tu2024stableanimator,tan2024animate}.
We select five challenging unseen scenes, including mirrored characters, celebrity portraits, and abstract characters.
We observe that almost all existing methods fail to generate satisfactory results across all reference examples.  
They struggle to generate accurate mirror content, preserve facial fidelity, and drive abstract human characters.
Although UniAnimate~\cite{wang2024unianimate} and Animate-X~\cite{tan2024animate} can maintain the identity, they fail to achieve pose-driven animation, making their outputs appear more like direct repetitions of the given images.
Furthermore, we provide more examples, including large human motions from jazz dance and higher dynamic actions, in Fig.~\ref{fig:more_visualization} and Fig.~\ref{fig:large_motion_cases}.

\subsection{Ablation Study}

\vspace{-5pt}
\begin{table*}[!ht]
\begin{minipage}[t]{0.5\linewidth}
\caption{PadaLN \textit{v.s.} UVT and Cross-attention.}
\centering
\small
\label{tab: pose_control_comparison}
\renewcommand{\arraystretch}{1.1}
\setlength{\tabcolsep}{3.5pt}
\resizebox{0.95\linewidth}{!}{
    \begin{tabular}{lcccccccc}
       \toprule
       \multirow{2}{*}{\textbf{Methods}} && \multicolumn{5}{c}{\textbf{Image Quality}} && \multicolumn{1}{c}{\textbf{Video}} \\
       \cmidrule{3-7} \cmidrule{9-9}
       && SSIM $\uparrow$ &&  PSNR* $\uparrow$ &&  LPIPS $\downarrow$ &&  FVD $\downarrow$ \\
       \midrule
       DynamiCtrl (\textit{w/} UVT) && 0.613 && 17.25 && 0.267 && 508.25 \\
       DynamiCtrl (\textit{w/} CA)  && 0.586  && 16.20  && 0.279 && 518.90 \\
       \midrule
       \textbf{DynamiCtrl (\textit{w/} PadaLN}) && \textbf{0.766} && \textbf{20.41}  && \textbf{0.166} && \textbf{152.31}  \\
       \bottomrule
    \end{tabular}}
\end{minipage}
\hfill
\begin{minipage}[t]{0.5\linewidth}
\centering
\small
\caption{Training with ``Joint-text'' paradigm.}
\label{tab: training_with_text_or_not}
\setlength{\tabcolsep}{3.5pt} 
\renewcommand{\arraystretch}{1.1}
\resizebox{0.95\linewidth}{!}{
    \begin{tabular}{lcccccccc}
       \toprule
       \multirow{2}{*}{\textbf{Methods}} && \multicolumn{5}{c}{\textbf{Image Quality}} && \multicolumn{1}{c}{\textbf{Video}} \\
       \cmidrule{3-7} \cmidrule{9-9}
       && SSIM $\uparrow$ &&  PSNR* $\uparrow$ &&  LPIPS $\downarrow$ &&  FVD $\downarrow$ \\
       \midrule
       DynamiCtrl$^\dag$ && 0.723 && 18.93 && 0.198 && 221.10 \\
       \textbf{DynamiCtrl}  && \textbf{0.766}  && \textbf{20.41}  && \textbf{0.166} && \textbf{152.31} \\
       \bottomrule
    \end{tabular}}
\end{minipage}
\vspace{-0.05in}
\end{table*}

\noindent \textbf{Effects of PadaLN control:} as shown in Table~\ref{tab: pose_control_comparison}, the proposed PadaLN demonstrates overall superior performance on TikTok, which is more suitable for the video DiT architecture. We believe that since PadaLN performs normalization on the pose tokens at each layer, followed by adaptive scaling and shifting, it adjusts the magnitude and distribution of pose features in a fine-grained manner within each full attention layer. This allows the transformer blocks to respond more smoothly to the control signals.
The results may also indicate that UVT and cross attention may not be the most effective mechanisms for injecting all kinds of control signals, especially in scenarios with sparse pose data and increasingly deep, parameter-heavy networks.

\noindent \textbf{Effects of ``Joint-text'' paradigm:}
as shown in Table~\ref{tab: training_with_text_or_not}, performing ``Non-text'' paradigm on a pre-trained model for this task can significantly affect its generative capability. Based on a 5B pre-trained model, we construct a baseline model with the ``Non-text'' paradigm (named ``DynamiCtrl$^\dag$''). Following methods~\cite{hu2024animate,zhang2024mimicmotion}, the baseline adopts a two-stage training strategy: first, fine-tuning the model after removing the text condition; then, introducing pose control by incorporating the Shared VAE encoder and the PadaLN. The results on TikTok demonstrate that the ``Joint-text'' paradigm yields a clear improvement, verifying the adverse effect of multi-stage training on the performance of pre-trained models in specific tasks. Therefore, we believe that, among the existing powerful diffusion transformer backbones, ``Joint-text'' paradigm is more suitable for this task.

\begin{table*}[ht!]
\begin{minipage}[t]{0.5\linewidth}
\caption{Comparison of different pose encoders.}
\centering
\small
\label{tab:pose_encoder}
\renewcommand{\arraystretch}{1.1}
\setlength{\tabcolsep}{3.5pt}
\resizebox{0.95\linewidth}{!}{
    \begin{tabular}{lcccccccc}
       \toprule
       \multirow{2}{*}{\textbf{Methods}} && \multicolumn{5}{c}{\textbf{Image Quality}} && \multicolumn{1}{c}{\textbf{Video}} \\
       \cmidrule{3-7} \cmidrule{9-9}
       && SSIM $\uparrow$ &&  PSNR* $\uparrow$ &&  LPIPS $\downarrow$ &&  FVD $\downarrow$ \\
       \midrule
       DynamiCtrl$^\dag$-Expert && 0.715 && 18.44 && 0.203 && 220.09 \\
       DynamiCtrl$^\dag$-Shared && 0.723 && 18.93 && 0.198 && 221.10 \\
       \midrule
       DynamiCtrl-Expert && 0.751 && \textbf{20.62} && 0.170 && 155.20 \\
       \textbf{DynamiCtrl-Shared}  && \textbf{0.766}  && 20.41 && \textbf{0.166} && \textbf{152.31} \\
       \bottomrule
    \end{tabular}}
\end{minipage}
\hfill
\begin{minipage}[t]{0.49\linewidth}
\centering
\figcaption{Effectiveness of aligned image and text.}
\includegraphics[width=\linewidth]{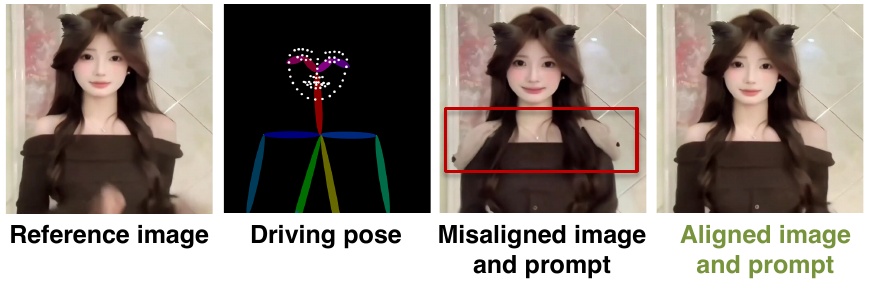}
\label{fig:aligned_image_text}
\end{minipage}
\vspace{-10pt}
\end{table*}

\noindent \textbf{Effects of shared VAE encoder:} compared to the expert pose encoder, using the VAE encoder already present in the diffusion architecture and requiring no additional training to encode pose signals is undoubtedly a more efficient approach. We validate the effectiveness of the shared VAE encoder on both the ``Non-text'' DynamiCtrl$^\dag$ and the ``Joint-text'' DynamiCtrl on TikTok, with the results recorded as ``x-Expert'' and ``x-Shared'', respectively. As shown in Table~\ref{tab:pose_encoder}, we observe that using the VAE encoder achieves better performances than the expert pose encoder. This effectively validates that there is no need to train an additional pose encoder for encoding pose sequences.

\noindent \textbf{Alignment between image and text conditions:}
we investigate the impact of a mismatched textual prompt and image on the model's generation quality. We use text generated by the VLM model~\cite{wang2024qwen2} as the aligned text, and construct misaligned text by modifying certain attributes. Since our training process tightly couples the semantics of the image and text, as shown in Fig.~\ref{fig:aligned_image_text}, mismatches between the conditions significantly degrade the generation quality.

\vspace{-5pt}
\begin{figure*}[ht]
    \centering
    \captionsetup{type=figure}
    \includegraphics[width=1.0\linewidth]{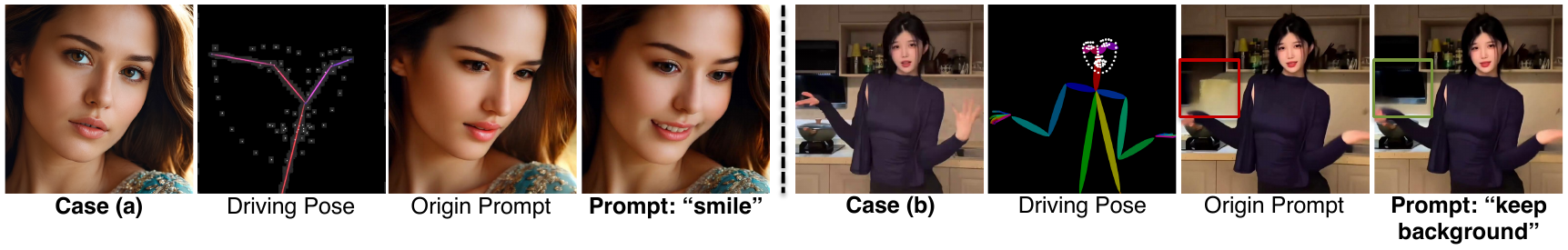}
    \vspace{-15pt}
    \caption{Owing to the effective text control, the quality of the videos can be readily improved.}
    \label{fig:text_control_ablation}
\end{figure*}
\vspace{-5pt}

\noindent \textbf{Positive effect of text guidance:}
as one of the advantages of the ``Joint-text'' paradigm, we further investigate the positive guidance effect of text, as illustrated in Fig.~\ref{fig:text_control_ablation}. In two cases, we enhance the original prompts provided by the VLM model~\cite{wang2024qwen2} with additional textual cues, enabling fine-grained control over human actions and mitigating temporal inconsistencies. In contrast, such effects are difficult to achieve under previous works~\cite{hu2024animate,zhang2024mimicmotion,tan2024animate,chang2023magicpose,xu2024magicanimate} in ``Non-text'' paradigm, especially in terms of detailed control, where one could only resort to randomly sampling a large number of seeds in hopes of obtaining satisfactory results. In comparison, our Joint-text paradigm facilitates more precise and efficient high-quality control in video generation.

\begin{wraptable}[6]{r}{0.49\textwidth}
\setlength{\tabcolsep}{0.7mm}
\centering
\small
{
\vspace{-30pt}
\caption{User study of human preferences with 12 volunteers evaluating 15 unseen data samples.}
\resizebox{1.0\linewidth}{!}{
\begin{tabular}{lcccccc}
   \toprule
   \textbf{Methods} && \textbf{ID preservation} $\uparrow$ &&  \textbf{Motion-align} $\uparrow$ &&  \textbf{Visual quality} $\uparrow$ \\
   \midrule
   MimicMotion~\cite{zhang2024mimicmotion} && 22.22 && 17.78 && 15.56 \\
   ControlNext~\cite{peng2024controlnext} && 22.22 && 14.44 && 18.89 \\
   StableAnimator~\cite{tu2024stableanimator}  && 14.45  && 21.11 && 17.78 \\
   Animate-X~\cite{tan2024animate}  && 14.44  && 22.23 && 14.44 \\
   \midrule
   \textbf{Ours}  && \textbf{26.67}  && \textbf{24.44} && \textbf{33.33} \\
   \bottomrule
\end{tabular}}
\label{tab: user_study}
}
\end{wraptable}

\subsection{User Study and More Applications}
In Table~\ref{tab: user_study}, we provide the results of the user study from a human visual perception perspective.
We evaluate three aspects: ID preservation, motion alignment between the driving video and the generated video, and overall video quality. Each metric accounts for 100\% in total, and the scores for each method are converted into percentages, making it easy to compare the performance of different methods across these evaluation metrics.

\vspace{-2pt}
\section{Applications} \label{sec:applications}
\begin{figure*}[ht!]
    \centering
    \captionsetup{type=figure}
    \includegraphics[width=0.92\linewidth]{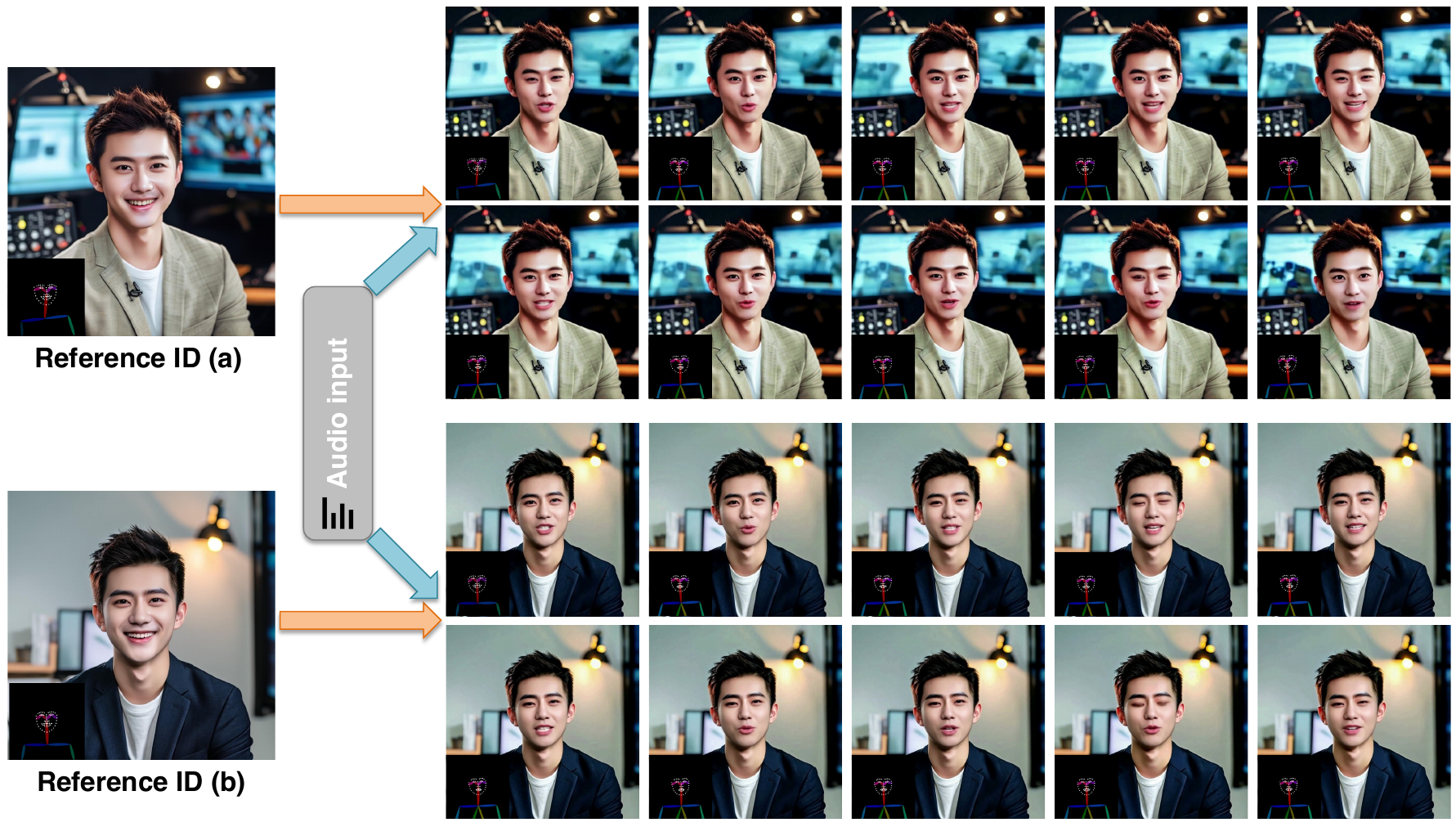}
    \caption{Visualization of digital humans. Given the digital human appearances (a) and (b), we can generate vivid digital human representations based on audio content.}
    \label{fig:digital_human}
\end{figure*}

\vspace{-2pt}
\noindent\textbf{Digital humans.}
\label{appendix:digital_human}
Due to the high-quality generation quality and highly robust identity preservation of our model, we explore its application in the field of digital humans. First, our model generates motion-capable character videos based on pose and textual prompts. It is important to note that since our model does not incorporate audio signals and has not been specifically trained on lip-sync datasets, it cannot directly drive character lip movements based on audio content. Subsequently, by leveraging the existing lip-sync model, we input the generated videos and audio signals into MuseTalk~\cite{zhang2024musetalk}, which ultimately synthesizes a vivid and lip-sync-consistent digital human representation.
Note that MuseTalk is only responsible for generating the mouth movements, while the overall character appearance and motion are still controlled by our DynamiCtrl model.
When generating digital humans, we first use a text-to-image model to create the initial frame, then generate videos of two different characters driven by the same pose video and audio clip, as shown in Fig.~\ref{fig:digital_human}.
It can be found that the digital humans we generate are not static but possess motion capabilities similar to those of real humans. Additionally, because the video quality generated in the first stage is high, MuseTalk is able to perform more effectively.

\begin{figure*}[ht!]
    \centering
    \captionsetup{type=figure}
    \includegraphics[width=1.0\linewidth]{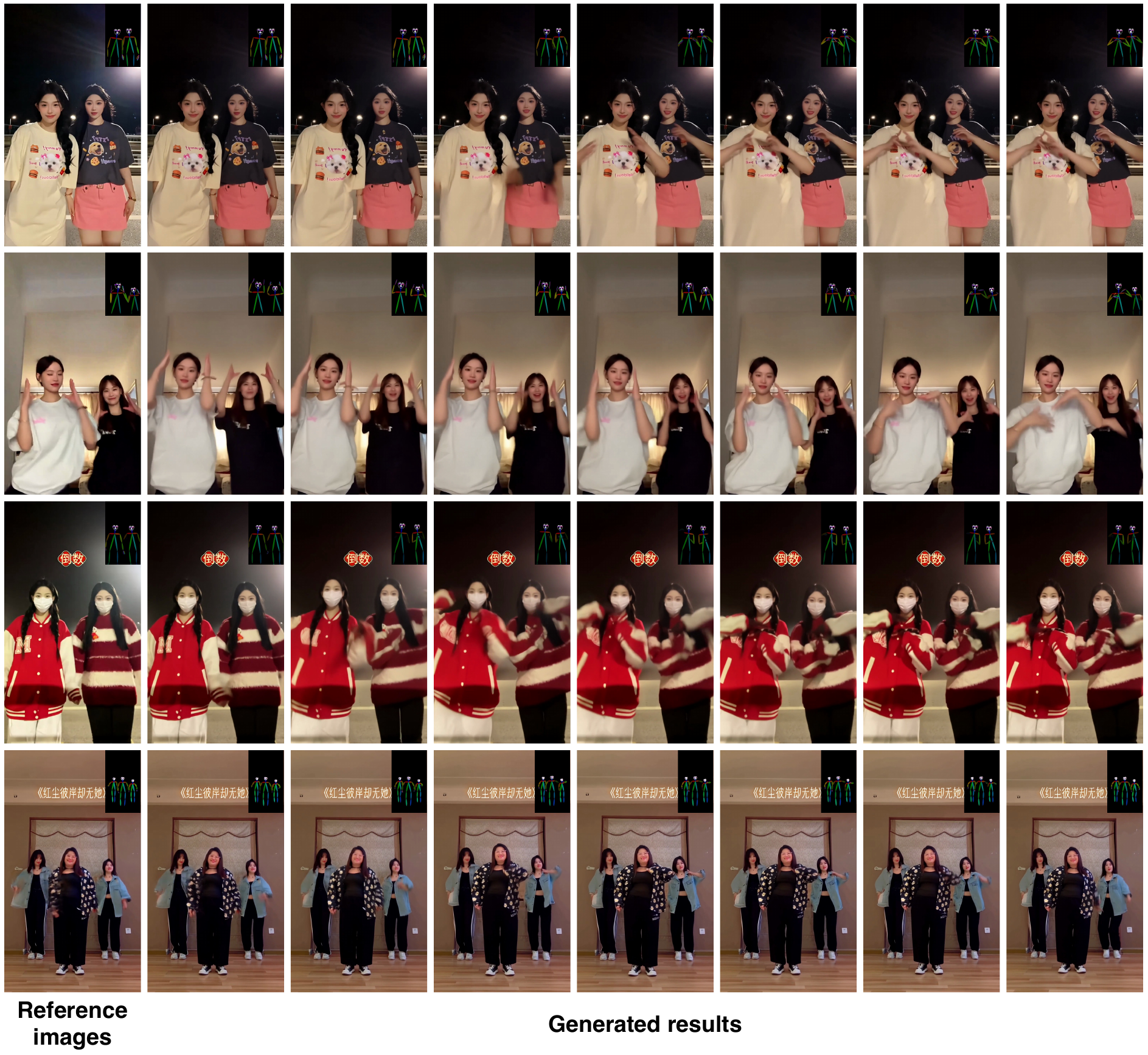}
    \caption{Human image animation in multi-person scenes. Our model can not only drive single-person scenes guided by driving pose video but also handle multi-person scenes.}
    \label{fig:multiple_person}
\end{figure*}

\noindent\textbf{Multiple person animations.}
\label{appendix:multiple_person}
In addition, we have explored video driving in multi-person scenarios. In multi-person scenarios, simultaneously driving the pose motions of multiple characters presents a significant challenge. We provide the results of our model in Fig.~\ref{fig:multiple_person}. The results show that our method can effectively control motions in such scenarios.

\vspace{-2pt}
\section{Implementation details} \label{sec:applications}
\vspace{-2pt}
The training of our DynamiCtrl is conducted on 8 NVIDIA H20 GPUs.
We build the whole framework and baseline models based on pre-trained CogVideoX~\cite{yang2024cogvideox} model.
We use the DWPose tool~\cite{yang2023effective} to estimate human pose, including body and hand parts. We employ the Qwen2-VL method~\cite{wang2024qwen2} to obtain image descriptions during both training and inference. To get the mask area of the objects in videos, we use the Segment-and-Tracking tool~\cite{cheng2023segment}.
Our model is trained with a batch size of 56 at a resolution of $1024 \times 576$ and 32 at a resolution of $1360 \times 768$, with a learning rate of $1e^{-5}$.

\textbf{Metrics.}
Following~\cite{wang2024disco,xu2024magicanimate,tan2024animate}, we adopt the L1, SSIM~\cite{wang2004image}, PSNR~\cite{hore2010image}, LPIPS~\cite{zhang2018unreasonable}, FID-VID~\cite{balaji2019conditional} and FVD~\cite{unterthiner2018towards} metrics for evaluation. Besides, the temporal consistency and text alignment based on feature similarity are also conducted for ablation.

\vspace{-2pt}
\section{Conclusion} \label{sec:exp}
\vspace{-2pt}
In this work, we proposed DynamiCtrl, a novel framework based on video DiT for pose-guided human image animation. Instead of an expert pose encoder, we leveraged the Shared VAE encoder for vision inputs. Our Pose-adaptive Layer Norm effectively injects pose features into attention blocks, enabling precise motion control. Additionally, we propose the ``Joint-text'' paradigm and achieved fine-grained video control. Extensive experiments confirmed that DynamiCtrl generates high-quality, identity-preserving, heterogeneous character driving, and highly controllable human motion videos.

\textbf{Limitations}
While video diffusion-based pose-guided human animation models, including the proposed DynamiCtrl framework, offer many advantages, several limitations must be considered. These models often require substantial computational resources and time to train and run, which can be a barrier for applications needing real-time processing or for users with limited access to high-performance computing resources.

\newpage
\bibliographystyle{IEEEtran}
\bibliography{neurips_2025}

\end{document}